\documentclass[10pt,twocolumn,letterpaper]{article}

\usepackage{cvpr}              % To produce the CAMERA-READY version
\usepackage[]{cvpr} % To force page numbers, e.g. for an arXiv version

\usepackage{graphicx}
\usepackage{amsmath}
\usepackage{amssymb}
\usepackage{booktabs}

\usepackage{times}
\usepackage{epsfig}
\usepackage{graphicx}
\usepackage{amsmath}
\usepackage{amssymb}

\usepackage{multicol}
\usepackage{multirow}
\usepackage{color}
\usepackage{makecell}
\usepackage{booktabs}
\usepackage{float}
\usepackage{dsfont}

\usepackage[pagebackref,breaklinks,colorlinks]{hyperref}

\usepackage[accsupp]{axessibility}  % Improves PDF readability for those with disabilities.

\usepackage[capitalize]{cleveref}
\crefname{section}{Sec.}{Secs.}
\Crefname{section}{Section}{Sections}
\Crefname{table}{Table}{Tables}
\crefname{table}{Tab.}{Tabs.}

\usepackage[font=small,labelfont=bf]{caption}  % set global caption size
\usepackage{makecell}
\usepackage{tabulary}
\definecolor{demphcolor}{RGB}{144,144,144}
\newcommand{\demph}[1]{\textcolor{demphcolor}{#1}}
\definecolor{mygray}{gray}{0.4}
\usepackage{pifont}% http://ctan.org/pkg/pifont for xmark and cmark
\newlength\savewidth

\newcommand{\tablestyle}[2]{\setlength{\tabcolsep}{#1}\renewcommand{\arraystretch}{#2}\centering\footnotesize}
\makeatletter\renewcommand\paragraph{\@startsection{paragraph}{4}{\z@}
  {.5em \@plus1ex \@minus.2ex}{-.5em}{\normalfont\normalsize\bfseries}}\makeatother
\def\x{$\times$} %% use as 2\x2 in text
\usepackage{xspace}

\usepackage{color, colortbl}
\definecolor{Gray}{gray}{0.9}

\definecolor{baselinecolor}{gray}{.9}

\begin{document}

\title{Accelerating Vision-Language Pretraining with Free Language Modeling}

\author{
Teng Wang$^{1,2\dagger}$, Yixiao Ge$^{3}$, Feng Zheng$^{1,5*}$, Ran Cheng$^1$, Ying Shan$^3$, Xiaohu Qie$^4$, Ping Luo$^{2,6}$ \\
$^1$Southern University of Science and Technology \ $^2$The University of Hong Kong \\ $^3$ARC Lab, $^4$Tencent PCG \ \ $^5$Peng Cheng Laboratory \ \ $^6$Shanghai AI Laboratory\\
{\tt\small tengwang@connect.hku.hk\ \  \{yixiaoge, yingsshan, tigerqie\}@tencent.com} \ \ \\ 
{\tt\small f.zheng@ieee.org\ \ ranchengcn@gmail.com\ \ pluo@cs.hku.hk}
}

\maketitle

%%%%%%%%% ABSTRACT
\begin{abstract}

\let\thefootnote\relax\footnotetext{$\dagger$ Work done during internship in ARC Lab, Tencent PCG. \\ {\hspace*{1.5em} $*$ Corresponding author}}

   The state of the arts in vision-language pretraining (VLP) achieves exemplary performance 
  but suffers from high training costs resulting from slow convergence and long training time, especially on large-scale web datasets. 
  An essential obstacle to training efficiency lies in the entangled prediction rate {\color{black}(percentage of tokens for reconstruction)} and corruption rate {\color{black}(percentage of corrupted  tokens)} in masked language modeling (MLM), 
  that is, a proper corruption rate is achieved at the cost of a large portion of output tokens being excluded from prediction loss.  
To accelerate the convergence of VLP, we propose a new pretraining task, namely, free language modeling (FLM), that enables a 100\% prediction rate with arbitrary corruption rates.
FLM successfully frees the prediction rate from the tie-up with the corruption rate while allowing the corruption spans to be customized for each token to be predicted.
FLM-trained models are encouraged to learn better and faster given the same GPU time by exploiting bidirectional contexts more flexibly.
Extensive experiments show FLM could achieve an impressive $2.5\times$ pretraining time reduction in comparison to the MLM-based methods, while keeping competitive performance on both vision-language understanding and generation tasks. Code will be public at~\url{https://github.com/TencentARC/FLM}.

\end{abstract}

\section{Introduction}

\begin{figure}
    \centering
    \includegraphics[width=0.46 \textwidth]{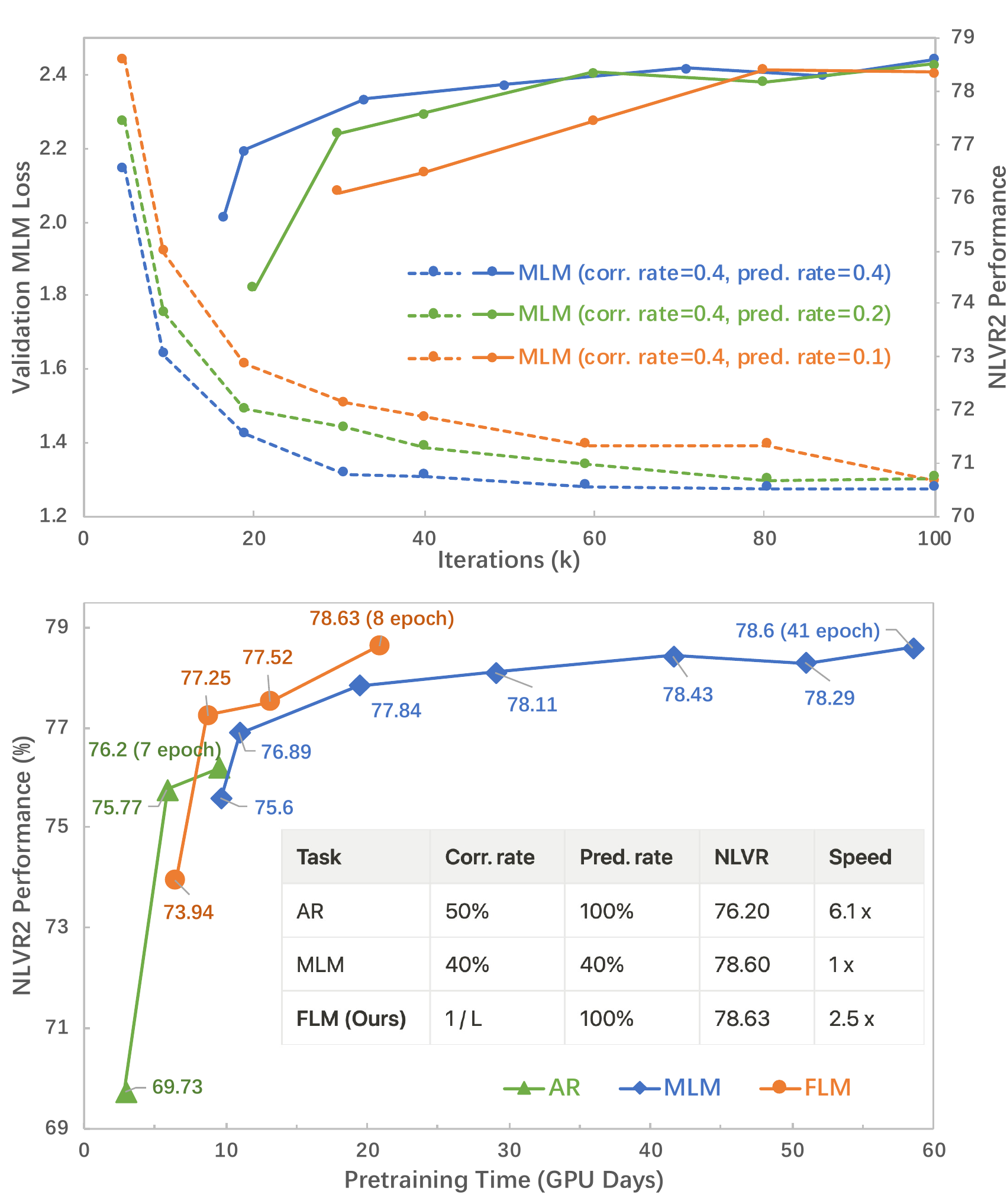}
    \vspace{-.5em}
    \caption{(a) Large prediction rate accelerates training. 
    Given a fixed corruption rate, we vary the prediction rate by randomly selecting a subset of output tokens for prediction loss. The learning rate schedule follows METER~\cite{dou2022empirical}. (b) The proposed FLM achieves competitive performance compared with MLM meanwhile significantly accelerating the pretraining stage. The downstream performance on NLVR$^2$~\cite{suhr2018corpus} is reported. 
    We show accuracy curves before convergence for better visualization. }
    \vspace{-1.0em}
    \label{fig:intro}
\end{figure}

Vision-language pretraining (VLP) has recently demonstrated impressive performance on a handful of vision-language tasks~\cite{li2019visualbert,chen2020uniter,li2020oscar,jia2021scaling,li2021align,dou2022empirical}, \eg, visual question answering, cross-modal retrieval, and image captioning. Several factors are responsible for the success: the availability of large-scale image-text datasets collected from the web~\cite{sharma2018conceptual}, high-capacity model architectures like Transformer~\cite{vaswani2017attention}, and effective pretraining objectives for cross-modal learning. 

One of the dominant pretraining objectives is masked language modeling (MLM), which was first introduced in natural language processing~\cite{devlin2018bert} and has been applied to vision-language areas in recent years~\cite{li2019visualbert}. MLM is a generative pretraining task designed to reconstruct a few (usually 40\% for VLP) masked text tokens via reasoning among the context of the remaining texts and the paired image. While effective in capturing cross-modal interactions, MLM-based methods~\cite{chen2020uniter,li2020unimo,kim2021vilt} suffer from slow convergence and long training time, especially for large-scale models and noisy web data.

We argue that the limited {prediction rate} in MLM impedes the convergence speed of pretraining, since a large portion of tokens accompanied by corruption are excluded from prediction loss.
As shown in Fig.~\ref{fig:intro}~(top), under the same {corruption rate} 
, a larger prediction rate for MLM results in faster convergence of validation loss and downstream performance.
It is intuitive to set a prediction rate of 100\% to fully exploit text tokens.
However, a paradox emerges where a large prediction rate can only be achieved with a greater corruption rate in MLM, but an extremely large corruption rate leads to an extremely tough pretraining task that may cause training collapse.

Autoregressive language modeling (AR) provides a workable solution to enable a 100\% prediction rate. It predicts the next token given the observation of previous tokens. As shown in Fig.~\ref{fig:intro} (bottom), AR performs favorably in training efficiency against MLM, \ie, 6.1\x speed-up for convergence. However, the converged performance by AR is, unfortunately, much inferior to MLM.
It is probably caused by the sub-optimal unidirectional corruption pattern, which is insufficient for downstream understanding tasks that usually rely on bidirectional contexts.

\begin{figure*}
    \centering
                \includegraphics[width=0.9 \textwidth]{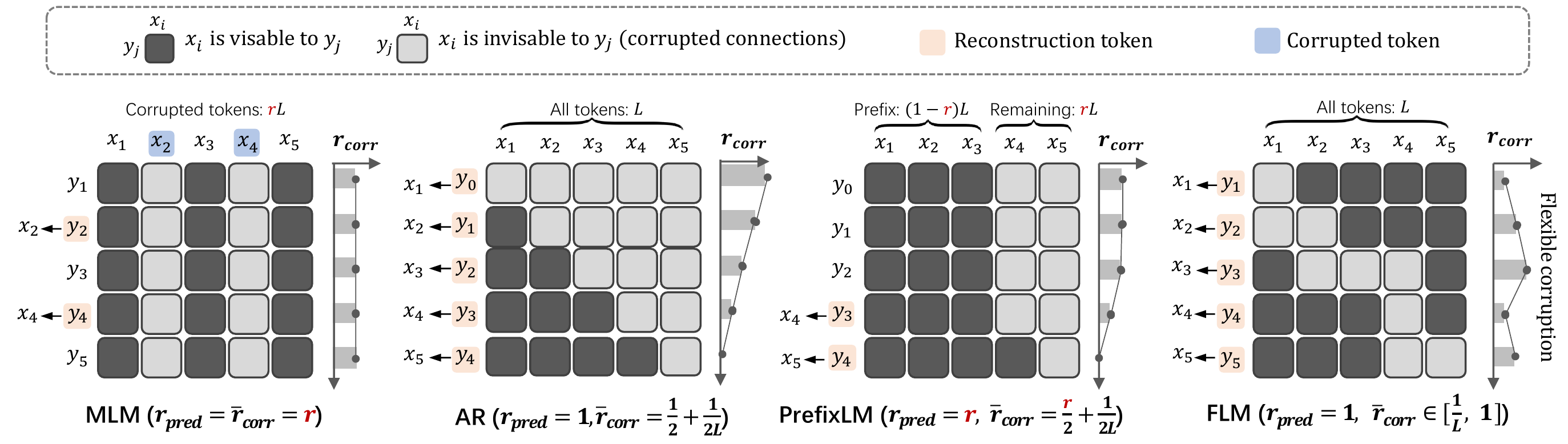}
    \vspace{-.5em}
    \caption{Dependency matrix of different language modeling methods in vision-language pretraining. $r_{\rm pred}$ represents the proportion of output tokens for reconstruction. $r_{\rm corr}$ represents the proportion of corrupted inputs for each output token. $\overline{r}_{\rm corr}$ is the mean corruption rate of all reconstruction tokens. FLM has distinct advantages compared with others: 1) Different from MLM and PrefixLM that bind  $r_{\rm pred}$ and $\overline{r}_{\rm corr}$ together by $r$, the unbound prediction rate in FLM could achieve 100\% for accelerating training as much as possible. 2) Without relying on a position-aware unidirectional corruption in AR/PrefixLM or fixed corruption across all positions in MLM (see the right-side line graph), the corrupted span in FLM for each output token could be different, and the corrupted rate is independent of the position of the output token, enabling a more flexible corruption pattern for better exploiting the bidirectional context information.} 
    \vspace{-1.0em}
    \label{fig:LMsComparison}
\end{figure*}

A natural question arises, \textit{can we accelerate the convergence of VLP by predicting 100\% tokens like AR meanwhile achieving competitive performance with MLM?} 
Towards this end, we introduce a new pretraining task, dubbed free language modeling (FLM), for VLP, that enjoys an extreme 100\% prediction rate and flexible bidirectional contextualized representations. 
We for the first time break up the entanglement between corruption and prediction rates, making the two factors freely determined. 
Furthermore, for each output token to be predicted, we allow independent and arbitrary-length spans (from one to 100\% tokens) as corrupted connections. 
Rather than the suffix-like corruption pattern as in AR (as well as PrefixLM~\cite{wang2021simvlm}),
the corruption span of FLM is primarily distributed in the middle of the sequence, establishing a flexible perception of bidirectional contexts for better adaptation to VL understanding tasks. The comparison between different pretraining objectives is illustrated in Fig.~\ref{fig:LMsComparison}.

To perform VLP with FLM, we propose an encode-corrupt-predict framework, which performs feature encoding once and reconstructs several corrupted versions of the text sequence in parallel. 
In the encoding step, bidirectional representations are achieved by learning forward and reverse unidirectional representations respectively, the order of which is manipulated by (reverse) casual masks in the same text Transformer.
Subsequently, we ensure a 100\% prediction rate by customizing corruption-prediction tasks for predicting each input token. 
In each corruption-prediction task, a span of corruption is randomly sampled and attached to the encoded sequence, followed by a reconstructor to solve the prediction task by reasoning among the remaining contexts. 
Unlike previous works (\eg, MLM, AR) that adopt pre-encoding corruption, we inject corruptions after one-time feature encoding, encouraging flexible corruption patterns and efficient parallel prediction.

Our contributions are three-fold.
(1) A novel pretraining objective for VLP, namely, free language modeling (FLM), is proposed to free the prediction rate from the constraints of corruption rate, enabling an appealing 100\% prediction rate for accelerating convergence speed during pretraining. 
(2) An encode-corrupt-predict framework built upon FLM objective is proposed, allowing efficient and effective learning of a set of prediction tasks by merely conducting feature encoding once.
(3) Extensive experiments on VQA, NLVR$^2$, image captioning, and image-text retrieval demonstrate the effectiveness of our FLM, where comparable performances to MLM are achieved with less than 50\% pretraining time.

\section{Related Work}

\paragraph{Vision-Language Pretraining.}
Vision-language pretraining tasks can be divided into two categories: (i) \textit{discriminative} tasks, \eg, image-text contrastive (ITC), image-text matching (ITM), and (ii) \textit{generative} tasks, \eg, masked language modeling (MLM) and autoregressive language modeling (AR). \textit{Discriminative} tasks consider the image-text pairs as multi-modal views of the same semantics. Contrastive or multi-view learning is adopted for learning the alignment between multiple modalities. For example, CLIP~\cite{radford2021learning}, ALIGN~\cite{jia2021scaling}, and following works~\cite{yao2021filip,li2022grounded,li2021align} utilize cross-modal contrastive learning by projecting the image and language information into a joint (structured) semantic space. \textit{Generative} tasks aim to reconstruct the corrupted text (image) with the assistance of visual (text) modality. The main body of representative works~\cite{li2019visualbert,lu2019vilbert,chen2020uniter,li2020oscar,zhang2021vinvl,wang2022vlmixer,zeng2021multi,bao2022vl}, 
employ the MLM-like objectives, where input texts(image) are partially masked and then interact with visual (text) tokens to reconstruct the corrupted part. SimVLM~\cite{wang2021simvlm} introduces a single prefix language modeling (PrefixLM) objective for exploiting large-scale weak supervision in VLP. CoCa~\cite{yu2022coca} further verifies the representation ability of autoregressive language modeling (AR) in the vision-language domain. 
While most existing methods combine discriminative and generative tasks for better representation learning, BEiT-3~\cite{wang2022image} shows a single generative language modeling (\eg, MLM) could handle the vision-language interactions and alignments well with the mixture-of-expert transformer. Although superior performance has been attained, most existing methods based on MLM suffer from low utilization of output tokens and lead to a slow convergence rate. This paper proposes a new generative language modeling method targeting pretraining acceleration. 

\vspace{-.2em}
\paragraph{Efficient Pretraining.}
While early VLP methods~\cite{tan-bansal-2019-lxmert,li2020oscar,chen2020uniter,zhang2021vinvl} rely on time-consuming pretrained object detectors for visual representation, PiexlBERT~\cite{huang2020pixel} and ViLT~\cite{kim2021vilt} directly apply grid/patch-level visual features to reduce computation complexity of the object-level visual encoder. Beyond the design of efficient model architecture, a few research focuses on data-efficient training. Bitton et al.~\cite{bitton2021data} propose an alternative masking strategy that better focuses on visually-related physical words to improve VLP in low-resource settings. DeCLIP~\cite{li2021supervision} enhances the CLIP by exploring more supervision signals, such as self-supervision within a single modality or multi-view supervision across different modalities. The most relevant work to this paper is GRIT-VLP~\cite{byun2022grit}, which assigns a larger mask rate for MLM and performs grouped in-batch negative sampling for ITC to accelerate the convergence. However, only half of the output tokens are assigned for the reconstruction task, where the under-used output tokens impede a further speed-up of pretraining. Our method decouples the corruption and reconstruction rate, making them freely chosen for a better combination between performance and efficiency. 

\vspace{-.2em}
\paragraph{Language Modeling.}
In NLP, MLM~\cite{devlin2018bert,liu2019roberta,he2020deberta} and AR~\cite{brown2020language,chowdhery2022palm} have been the two most popular generative pretraining objectives. AR aims to estimate the probability distribution of a given text sequence using the product rule by an auto-regressive model. However, unidirectional encoding may not be suitable for language understanding tasks that prefer bidirectional context information. MLM enables bidirectional contexts for language understanding tasks but can not be directly adopted into language generation tasks. Some works~\cite{yang2019xlnet,du2022glm} unify MLM and AR for better performance on both language understanding and generation tasks. Wettig et al.~\cite{wettig2022should} study the choice of the mask ratio in MLM from the perspective of both corruption and prediction. However, among previous methods, little attention has ever been devoted to the issue of training efficiency. We target accelerating vision-language pretraining meanwhile keeping decent performances on vision-language understanding and generation tasks.

\section{Method}
In this section, we first recap the representative language modeling methods for VLP from a corruption-prediction view in Sec.~\ref{sec:lmAsCP}. Then we propose the new language modeling method FLM to decouple the prediction rate from the corruption rate in Sec.~\ref{sec:flm}. Finally, we introduce FLM into VLP and propose a novel encode-corrupt-predict framework for accelerating VLP in Sec.~\ref{sec:VLPflm}.
\subsection{Language Modeling as Corruption-Prediction}
\label{sec:lmAsCP}
Given an input sequence $\mathbf{x}=\{x_1,...,x_L\}$, \textbf{MLM} aims to learn a deep bidirectional representation by randomly replacing part of input tokens with a special mask token, and then maximize the probability of reconstructing those masked tokens $P(\mathbf{x_m} | \mathbf{x_{\setminus m}})$, where $\mathbf{x_{ m}}$ represents corrupted tokens. 
\textbf{AR} uses a left-to-right autoregressive factorization to model the density distribution of the sequence $\sum_{i=1:L} \log P(x_{i}|x_{<i})$. 
\textbf{PrefixLM} enables bidirectional perception between prefix tokens and left-to-right autoregressive factorization to model the density distribution of the remaining sequence $\sum_{i=L_{p}:L} \log P({x}_{i} | {x}_{[L_p, i]}, {x}_{<L_p})$ where $L_{p}$ represents the prefix length.

Note that all the above methods could be interpreted as a corruption-prediction problem because each prediction token only has a partial observation of the input data, \ie, the interactions between output tokens and some input tokens are corrupted. Therefore, their pretraining objectives could be unified as maximizing the reconstruction probability:
\begin{equation}
% \vspace{-.3em}
  \begin{aligned}
\mathbb{E}_{\mathbf{M} \sim B(r)} \!\sum_{i  = 1:L}\! \mathds{1}_{m_{ii}=0} \log {P}(x_i | \{x_j| {{m}_{ij}} \!=\! 1\}), 
\end{aligned}
\label{eqn:reconst}
% \vspace{-.3em}
\end{equation}
where $\mathbf{M}=\left[ m_{ij} \right]_{1 \leq i \leq L, 1\leq j \leq L}$ represents the dependency matrix between the input and the prediction target, ${m_{ij}}=1/0$ represents that $x_j$ is visible/invisible when predicting ${x_i}$. $B(r)$ represents a distribution parameterized by $r$, which is customized by specific models. The dependency matrices $\mathbf{M}$ of different language modeling methods are as follows (also illustrated in Fig.~\ref{fig:LMsComparison}): 
\begin{itemize}
\setlength\itemsep{0em}
    \item \noindent For MLM, $\mathbf{m}_{1,:}\!=\!\cdots\!=\!\mathbf{m}_{L,:}\!=\!\mathbf{p} \!\sim\! \text{Binomial}(r_{\rm mask})$. The corruption for predicting all $x_i$ is the same and it is sampled from a Binomial distribution.
\item For LM, ${m_{ij}} \!=\! \mathds{1}_{j < i}$. The corruption for predicting $x_i$ depends on the position $i$, which gets shorter with a larger $i$.
\item For PrefixLM, $m_{ij}\!=\!\mathds{1}_{j < \max (i, L_{\rm p})}$, where prefix length $L_{\rm p} \!=\! (1-r_{\rm span})\cdot L$ and $r_{\rm span} \! \sim \! {\rm Uniform}(0,1)$ represents the length ratio of the corrupted span.
\end{itemize}

\subsection{Free Language Modeling (FLM)}
\label{sec:flm}
From the above analysis, the representative MLM or AR/PrefixLM methods have limited freedom of the dependency matrix, which is prone to the following issues: 1) the tie-up between the prediction and corruption rate in MLM may lead to a low convergence speed during training; 2) inflexible and non-customizable contexts for each prediction target result in sub-optimal context learning. For example, the suffix-like corruption in AR disables the bidirectional context modeling which is essential for downstream understanding tasks. Moreover, the autoregressive prior in AR results in uneven distribution of corruption rate. Latter tokens are always assigned with a smaller corruption rate, thus being easy-predictable compared with former ones. The position-related distribution of the corruption rate may cause a sub-optimal difficulty degree of pretraining tasks.

The goal of FLM is to disentangle the prediction and corruption rate for fully utilizing training signals to accelerate VLP. The model after disengagement has a more flexible corruption pattern that benefits bidirectional contextual representation learning. Following the unified formulation of the corruption-prediction problem in Eqn.~\ref{eqn:reconst}, we introduce the dependency matrix of FLM:
\begin{equation}
  \begin{aligned}
  {m_{ij}} = 1 \text{ if } m_{ij} \notin {\text{span}_i}, \text{otherwise }0.
\end{aligned}
\label{eq2}
\end{equation}
where $\text{span}_i$ is random span corruption with length $L^i_{\rm span}$ that satisfies $i \in \text{span}_i$. The starting position and length of $\text{span}_i$ could be customized or randomly sampled from a distribution. In our implementation, we sample $L^i_{\rm span} \!\sim\! {\rm Bernoulli}(L, r_{\rm corr})$ for each $i$, $r_{\rm corr}$ is the hyperparameter indicating the expected corruption rate.

Note that the corrupted span in FLM could differ for different predictions, hopefully increasing the flexibility of bidirectional contextualized interactions and helping optimization. Since the choice of $\text{span}_i$ does not interfere with each other for different $i$, the prediction rate could increase to 100\% which allows all input tokens to be reconstructed. As for the corruption rate, we note that the minimal corruption rate is $1/L$, since at least one token should be corrupted to reconstruct itself to avoid information leakage.

\begin{figure}
    \centering
    \includegraphics[width=0.5\textwidth]{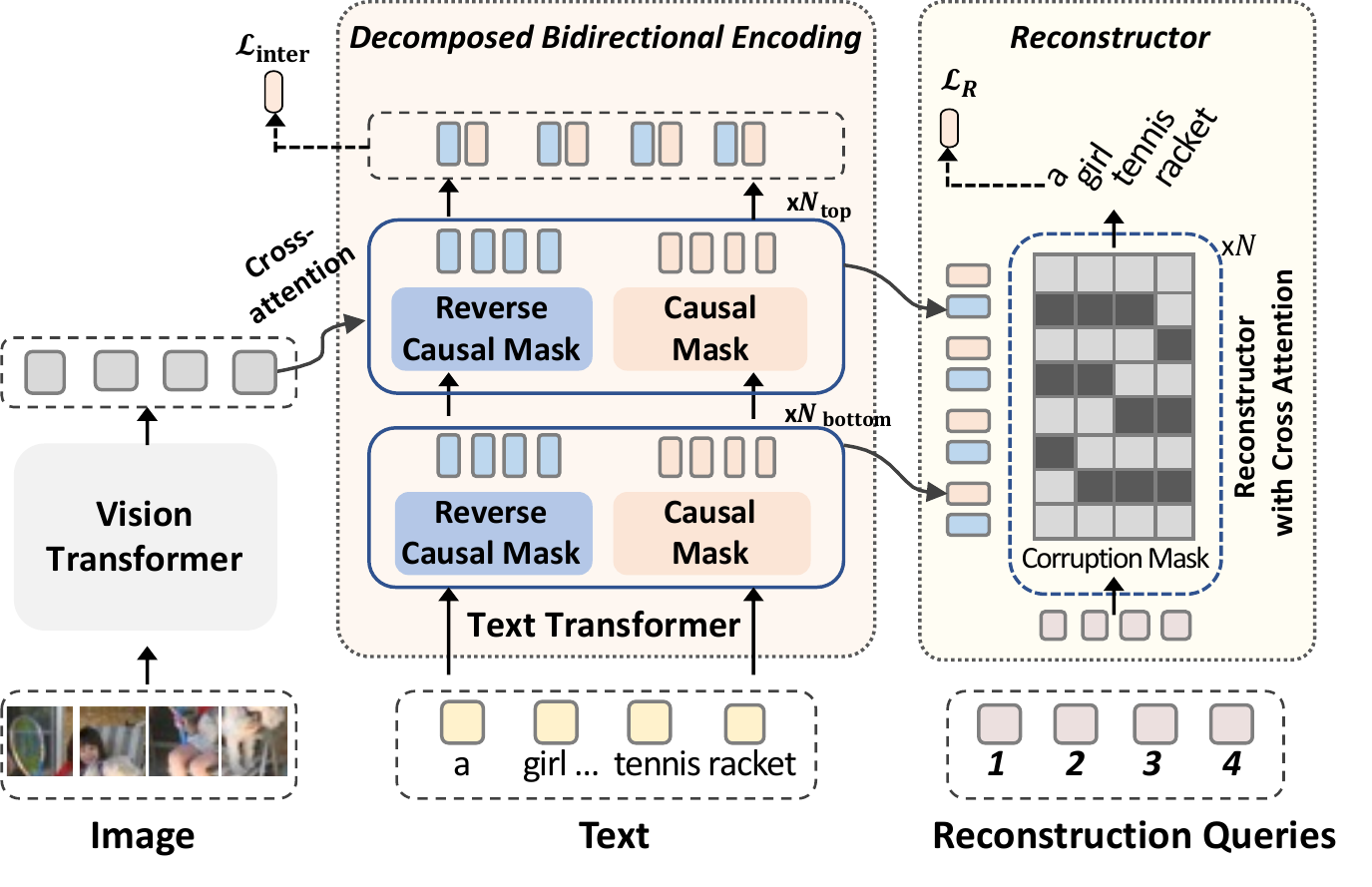}
    \vspace{-1em}
    \caption{Overview of the proposed VLP framework with free language modeling (FLM). First,
    the image is patchified and encoded by a vision transformer into a sequence of vision tokens. Then, the text transformer performs uni-modal text feature encoding in the bottom layers and multimodal fusion between visual and text features in the top layers. Bidirectional multimodal representations are achieved by learning forward and reverse unidirectional representations, respectively, the order of which is manipulated by (reverse) causal masks in the same text transformer. After feature encoding, we construct a set of independent corruption-prediction tasks. For each task, we inject a random span corruption into the multimodal representation and then introduce a reconstruction query that gathers informative contexts from the corrupted features for reconstructing a single target token. Benefiting from the flexibility of post-encoding corruption, 100\% text tokens could be efficiently reconstructed in parallel. 
    }
    \vspace{-1.0em}
    \label{fig:overview}
\end{figure}

\subsection{Vision-Language Pretraining with FLM}
\label{sec:VLPflm}

Built upon FLM, a new encode-corrupt-predict pretraining framework is proposed for efficient pretraining with decoupled prediction and corruption rates. Given the input sequence $\mathbf{x}=\{x_1,...,x_L\}$, we formulate the reconstruction of all input tokens as $L$  independent corruption-prediction tasks. For $i$-th task, the learning objective is to maximize ${P}(x_i | \{x_j| {{m}_{ij}} \!=\! 1\})$ by reasoning upon uncorrupted bidirectional tokens. Fig.~\ref{fig:overview} depicts the pipeline of the model.

\vspace{-.3em}
\paragraph{Decomposed Bidirectional Encoding.} 
Since FLM establishes a customized corruption span for each prediction task, 
a naive solution of repeating the MLM-style feature encoding (transformer with fully-visible attention mask) for each task becomes time-consuming. Instead, 
we propose to share the intermediate features of different tasks for efficient inference by decomposing the token representation into two complementary views, left-to-right, and right-to-left features. The former adopted a text transformer with a casual mask, enforcing the latter tokens attending on previous tokens. The latter adopted the same transformer with a reverse causal mask for a right-to-left context flow.

Specifically, we first encode image features by a CLIP image transformer.
For the text input, the $N_{\rm bottom}$ bottom layers of the text transformer perform the decomposed bidirectional encoding with the language input only, while $N_{\rm top}$ top layers further receive image features and fuse the multimodal features by cross-attention layers. 
After encoding, we obtain the bidirectional representation in text transformer, denoted as $E^{n}=\{e^{n}_1,\cdots, e^{n}_{L}\}$, where $e_i=\{e^{{\rm l2r}, n}_i, e^{{\rm r2l}, n}_i\}$ is the token representation for $x_i$ comprised of features from forward and reverse flow at the $n$-th layer.

\vspace{-.3em}
\paragraph{Reconstructor.} Following the description in Sec.~\ref{sec:flm}, we sample a dependency matrix $\mathbf{M}$ to construct several corruption-prediction tasks. As a consequence of the span corruption, some elements in $E^n$ that rely on corrupted inputs need to be neglected to avoid information leakage. 
To reconstruct $x_i$, we gather context from uncorrupted features in $E$ by cross-attention: 
\begin{equation}
  \begin{aligned}
    &q^{n+1}_i\!=\!\text{CrossAttention}(q^{n}_{i}, E^{n}_i) \\
    &E^{n}_i\!=\!\{e^{{\rm l2r}, n}_j | m_{ij}\!=\!1 \text{ \!,\! } j\!<\!i\} \!\cup\! \{e^{{\rm r2l},n}_j | m_{ij}\!=\!1 \text{ \!,\! } j\!>\!i\},
\end{aligned}
\label{eq2}
\end{equation}
where $E^{n}_i$ represents all uncorrupted elements in $E^{n}$ given $M$. $q_i$ is a learnable reconstruction query, which is initialized as the $i$-th positional embedding in the first layer. The selection process from $E^{n}$ to $E^{n}_i$ is implemented as a specific attention mask in cross-attention layers, as illustrated in Fig.~\ref{fig:overview}. 
By forwarding on stacked cross-attention layers, $q^n_{i}$ aggregates deep bidirectional contexts for effective reconstruction of $x_i$. The output features of the last layer in the reconstructor are input into an MLP for final prediction. 

Note that $q^n_i$ works independently from each other, making a flexible addition or removal of some tasks. By sharing feature encoding, all reconstruction tasks run in parallel with low computation consumption.

\vspace{-.3em}
\paragraph{Pretraining Objectives.}
The reconstruction objective is to minimize the negative log-likelihood of predicted tokens: 
\begin{equation}
  \begin{aligned}
\mathcal{L}_{\rm R} \!=\! -\mathbb{E}_{\mathbf{M} \sim B(r)} \!\sum_{i  = 1:L}\! \log {P}(x_i | \{x_j| {{m}_{ij}} \!=\! 1\}),
\end{aligned}
\label{eq2}
\end{equation}
To further enhance the model's representability, we introduce an intermediate prediction loss upon the $E^N$ that improves the local temporal dependency between words, where N represents the last layer of the text transformer. We supervised the forward/reverse sequence $e^{\rm l2r, N}_i$/$e^{\rm r2l, N}_i$ by their next/previous tokens. The intermediate loss is the summation of two unidirectional prediction problems:  $\mathcal{L}_{\text{inter}} = \mathcal{L}_{\rm l2r} + \mathcal{L}_{\rm r2l} = \sum_{i=1:L} \log P(x_{i}|x_{<i}) + \sum_{i=1:L} \log P(x_{i}|x_{>i})$. 
The overall pretraining objective of FLM is calculated by $\mathcal{L}_{\rm FLM} = \mathcal{L}_{\rm R} + \mathcal{L}_{\rm inter}$.

% :
% \begin{equation}
%     \mathcal{L}_{\rm FLM} = \mathcal{L}_{\rm R} + \mathcal{L}_{\rm inter}.
%     \label{eqn:sam}
% \end{equation}
% \vspace{-.5em}
\begin{table*}[]
    \small
    \centering
    \setlength{\tabcolsep}{1.6 mm}{
    \begin{tabular}{lll|cccccccc|c}
    \toprule
    \multirow{2}{*}{ Method}  & \multirow{2}{*}{$r_{\rm corr}$} &  \multirow{2}{*}{$r_{\rm pred}$} & VQA &  \multicolumn{2}{c}{NLVR$^2$} & \multicolumn{2}{c}{Retrieval (Flickr30K)} & \multicolumn{3}{c|}{COCO Captioning} & \multirow{2}{*}{GPU Days (speed-up)}\\
    & & & test-dev & dev & test & IR@1 & TR@1 & BLEU & METER & CIDEr & \\
    \midrule
% =======================
%AR
        AR & 50\% & $100\%$ & 72.85 & 75.79 & 76.29 & 66.59 & 84.10 & \underline{35.70} & \underline{28.86} & \underline{120.6} & \ \ 9.6 (6.1\x)\\
% =======================
%PrefixLM + GLM
        PrefixLM & $25\%$ & $50\%$ & 72.64 & 75.73 & 76.17 & 66.21 & 82.70 & 35.50 & 28.79 & 119.4 & 10.0 (5.9\x)   \\
% =======================
%MLM
        MLM & $15\%$ & $15\%$ & 73.52 & 77.46 & 78.28 & 71.33 & \underline{88.40} & 34.90 & 28.50 &  117.5 &   58.7 (1\x)\\
        MLM & $40\%$ & $40\%$ & \textbf{73.95} & \underline{77.62} & \underline{78.60} & \textbf{73.41} & \textbf{89.20} & 35.50 & 28.79 & 120.3 & 58.7 (1\x) \\
% =======================
%FLM
        \rowcolor{Gray}
        FLM (Ours) & $1/L$ & $100\%$ & \underline{73.85} & \textbf{77.99} & \textbf{78.63} & \underline{72.81} & 87.40 & \textbf{36.68} & \textbf{29.17} & \textbf{123.0}  & 22.7 (2.5\x) \\
\bottomrule
    \end{tabular}
    }
    \vspace{-.5em}    
    \caption{Performance Comparison between different language modeling methods. $r_{\rm corr}$ and $r_{\rm pred}$ refer to the corruption and prediction rates. 
    All models are based on CLIP-B/32 image encoder and a text transformer initialized by RoBERTa. Note that the default $r_{\rm corr}$ of FLM is set to $1/L$ for better efficiency, while FLM's performance will be further improved by an optimal $r_{\rm corr}$, as indicated in Table~\ref{tab:ablcr}.}
    \label{tab:lmcomparison}
    \vspace{-1.5em}
\end{table*}

\begin{table}[]
    \centering
    \small
    \setlength{\tabcolsep}{1.5 mm}{
    \begin{tabular}{l|ccccl}
    \toprule
    \multirow{2}{*}{Method} & VQA & \multicolumn{2}{c}{NLVR$^2$} & Captioning & \multirow{2}{*}{GPU Days}\\
    & test-dev & dev & test & CIDEr & \\
    \midrule
        \multicolumn{5}{l}{\it{\textbf{CLIP-B/32 on 13M data}}} \\
        AR & 73.46 & 76.60 & 77.21 & 121.5 & 21.3 (5.4\x)  \\
        MLM & 74.25 & 78.63 & 79.19 & {122.6} & 116.0 (1\x)\\
        \rowcolor{Gray}
        FLM & \textbf{74.28} & \textbf{78.73} & \textbf{79.52} & \textbf{122.6} & {\color{black}32.0 (3.6\x)} \\
        \multicolumn{5}{l}{\it{\textbf{CLIP-B/16 on 4M data}}} \\
        AR & 75.05 & 77.38 & 78.79 & 126.0 & 12.3 (5.0\x) \\ 
        MLM & 75.76 & \textbf{79.93} & 79.83 & 125.4 & 61.4 (1\x) \\ 
        \rowcolor{Gray}
        FLM & \textbf{75.95} & 79.02 & \textbf{80.03} & \textbf{126.5} & 16.1 (3.8\x)\\ 
\bottomrule
    \end{tabular}
    }
    \vspace{-.5em}
    \caption{Performance comparison of different pretraining objectives with a larger data scale (from 4M to 13M) or a larger number of patches (patch size from 32 to 16). 
    For 13M data, we extend the training iteration of MLM to 200k.
    }
    \vspace{-1.5em}
    \label{tab:datascale}
\end{table}

% \vspace{-1em}

\section{Experiments}
\subsection{Experimental Setting}
\paragraph{Pretraining Data.}
Following previous work~\cite{li2021align,dou2022empirical}, the pretraining data comes from four commonly used datasets, including COCO~\cite{lin2014microsoft}, Visual Genome~\cite{krishna2017visual}, SBU Captions~\cite{ordonez2011im2text}, and Conceptual Captions 3M~\cite{sharma2018conceptual}, totally 4M images. An enlarged version of $\sim$13M images is further used to boost performance by including Conceptual Caption 12M~\cite{changpinyo2021conceptual}~\footnote{Only ~8.7M images of CC12M are accessible due to expired URLs.}.

\vspace{-.3em}
\paragraph{Downstream Tasks.}
During finetuning, we append a special $\rm [CLS]$ token into the reconstructor and use its output features as the global cross-modal representation. We follow~\cite{chen2020uniter} to adapt the pre-trained model to four downstream vision-language understanding tasks, visual question answering~(VQA)~\cite{antol2015vqa}, natural language for visual reasoning~(NLVR$^2$)~\cite{suhr2018corpus}, image-text retrieval~(TR), text-image retrieval~(IR). We also test the performance on vision-language generation tasks, \ie, image captioning~\cite{lin2014microsoft}. For image captioning, we drop the reconstructor and use the text transformer with a causal mask for sequence generation. More details can be found in supplementary materials.

\vspace{-.3em}
\paragraph{Pretraining Details.}
Following ALBEF~\cite{li2021align} and METER~\cite{dou2022empirical}, the visual transformer is initialized by CLIP-ViT~\cite{radford2021learning} pretrained on 400M noisy image-text pairs. The visual transformer with ViT-B/32 is used as our base architecture for ablation study, and the one with ViT-L/14 is for scaling up to compare with other methods. We denote models with ViT-B image encoder as {Ours} and ViT-L as {Ours}$_{\rm LARGE}$. 
The bottom six layers of the text transformer are initialized by the bottom six layers of RoBERTa$_{\rm BASE}$~\cite{liu2019roberta}. The reconstructor is implemented by a 12-layer ($N_{\rm bottom}=N_{\rm top}=6$) transformer decoder (remove self-attention layers and only keep cross-attention layers) with a hidden size of 768 and a head number of 12. The default corruption rate of FLM is $1/L$, \ie, in each corruption-prediction task, only a single token is corrupted and then reconstructed from their contexts. While minimal corruption achieves decent performance, we further explore the choice of corruption rates in Sec.~\ref{subsec:label}. 

We pretrain the model for a maximum of 30k steps, with a total batch size of 4096 on 16 TITAN V100 GPUs by AdamW~\cite{loshchilov2017decoupled} optimizer and gradient accumulation. Mixed-precision training is used to reduce memory consumption and accelerate training. We use a 5\% warm-up schedule with a maximum learning rate of 4e-4. Following~\cite{dou2022empirical}, we assign a lower learning rate of 8e-5 for all pretrained layers. 
The text sequence length is limited to 50 subwords. More details are in the supplementary materials.

\vspace{-.3em}
\paragraph{Baselines.}
We also pretrain some generative language modeling methods for comparison, including MLM, AR, and PrefixLM~\cite{wang2021simvlm}. Specifically, we directly input the (corrupted) text sequence into the text transformer and then build an MLP layer upon the last layer of the text transformer for reconstructing the original input. For AR and PrefixLM, we follow the same learning rate schedule as FLM. For MLM, we follow~\cite{dou2022empirical} to train the model for 100k iterations with a maximum learning rate of 5e-5 and a warm-up rate of 10\%. To compare the convergence speed of different methods, we report the GPU days when reaching the best validation performance (\ie, reconstruction accuracy on the COCO validation set).

% overall table of all ablations
\begin{table*}[t]
% \vspace{-.2em}
\centering
\small
%#################################################
% Loss term
%#################################################
\subfloat[
\textbf{Loss term}. Intermediate losses are effective and complementary to FLM loss.
    \label{tab:auxLoss}
]{
% \centering
\begin{minipage}{0.29\linewidth}{\begin{center}
\renewcommand\arraystretch{1.0}
\setlength{\tabcolsep}{2.5 mm}{
    \begin{tabular}{c|cc}
    \toprule
        Loss & VQA & NLVR$^2$ \\
        \midrule
        $L_{R}$ & 73.04 & 77.18 \\
        $L_{R} + L_{\rm r2l}$ & 73.38 & 77.59 \\
        $L_{R} + L_{\rm l2r}$ & 73.67 & 78.00 \\
        \rowcolor{Gray}
        $L_{R} + L_{\rm inter}$  & \textbf{73.85} & \textbf{78.63} \\
    \bottomrule
    \end{tabular}}
    \label{tab:auxLoss}
\end{center}}\end{minipage}
}
\hspace{2em}
%#################################################
% Parameter sharing
%#################################################
\subfloat[
\textbf{Parameter sharing}.
Sharing two uni-directional encoders is effective and efficient.
        \label{tab:sharing}
]{
\begin{minipage}{0.29\linewidth}{\begin{center}
\renewcommand\arraystretch{1.66}
% \tablestyle{10pt}{1.05}
\setlength{\tabcolsep}{0.95 mm}{
    \begin{tabular}{c|cc}
    \toprule
        Text Encoder & VQA & NLVR$^2$ \\
        \midrule
        Unshared text encoder & 73.46 & 77.51 \\
        \rowcolor{Gray}
        Shared text encoder & \textbf{73.85} & \textbf{78.63} \\
    \bottomrule
    \end{tabular}}
\end{center}}\end{minipage}
}
\hspace{2em}
%#################################################
% Prediction rate
%#################################################
\subfloat[
\textbf{Prediction rate.} FLM with a larger prediction rate improves performance.
    \label{tab:ablpr}
]{
\begin{minipage}{0.29\linewidth}{\begin{center}
\renewcommand\arraystretch{1.0}
% \tablestyle{10pt}{1.05}
\setlength{\tabcolsep}{2.2 mm}{
% \vspace{-1.2em}
\begin{tabular}{c|cc}
\toprule
    Prediction rate & VQA & NLVR$^2$   \\  
    \midrule
    50\% & 73.74 & 77.47  \\
    75\% & 73.89 & 77.65  \\
    90\% & \textbf{74.00} & 78.17  \\
    \rowcolor{Gray}
    100\% & 73.85 & \textbf{78.63} \\
\bottomrule
\end{tabular}}

\end{center}}\end{minipage}
}
\\
\centering
\vspace{-.5em}
%#################################################
% Corruption Rate
%#################################################
\subfloat[
\textbf{Corruption Rate}. FLM enables a flexible choice of the corruption rate.
    \label{tab:ablcr}
]{
\begin{minipage}{0.35\linewidth}{\begin{center}
\renewcommand\arraystretch{1.0}
\setlength{\tabcolsep}{2.0 mm}{
    \begin{tabular}{l|cc}
    \toprule
        Corruption & VQA & NLVR$^2$ \\
        \midrule
        \rowcolor{Gray}
        span corruption ($1/L$) & {73.85} & {78.63} \\
        span corruption (30\%)  & {73.96} & \textbf{78.83}  \\
        span corruption (40\%)  & \textbf{74.04} & 78.82  \\
        span corruption (50\%)  & 74.01  & 77.84  \\
        random corruption (15\%) & 73.93 & 78.38 \\
        random corruption (30\%) & 73.69 & 77.74 \\
    \bottomrule
    \end{tabular}}
\end{center}}\end{minipage}
}
\hspace{4em}
%#################################################
% Number of reconstruction Layers
%#################################################
\subfloat[
\textbf{Number of reconstruction Layers}. FLM benefits from a deeper reconstructor.
\label{tab:layerNum}
]{
\centering
\begin{minipage}{0.35\linewidth}{\begin{center}
\renewcommand\arraystretch{1.17}
\setlength{\tabcolsep}{4.0 mm}{
% \vspace{-2em}
    \begin{tabular}{cc | cc}
    \toprule
        Bottom & Top & VQA & NLVR$^2$ \\
        \midrule
        $\times$  & 1 & 73.46 & 77.49 \\
        $\times$  & 3 & 73.62 & 78.20 \\
        $\times$  & 6 & 73.74 & {78.14} \\
        3 & 6 & 73.69 & 78.20 \\
        \rowcolor{Gray}
        6 & 6 & \textbf{73.85} & \textbf{78.63} \\
    \bottomrule
    \end{tabular}}
\end{center}}\end{minipage}
}
\vspace{-.5em}
\caption{
FLM ablation experiments with ViT-B/32 pretrained on 4M data. We report the finetuned accuracy (\%) on the VQA test-dev and NLVR$^2$ test set. 
Default settings are marked in \colorbox{baselinecolor}{gray}.
}
\vspace{-1.5em}
\label{tab:ablations} 
\end{table*}

\subsection{Comparison with Language Modeling Methods}

As shown in Table~\ref{tab:lmcomparison}, compared with MLM, the proposed FLM achieves a 2.5\x speed-up while keeping the comparable performance on VL understanding tasks and superior performance on VL generation tasks. 

AR achieves decent performance on image captioning generation but inferior performance on other VL understanding tasks, due to the lack of ability to capture bidirectional interactions among sequences. Moreover, AR has a faster convergence rate and high training efficiency. Although PrefixLM enables bidirectional interactions between partial inputs, which is beneficial for VL classification tasks, the performance of PrefixLM is similar to AR. However, the reconstruction target mainly falls on the right side of the sequence, where the uneven distribution may push the learned representation towards an unsatisfactory language prior. For MLM, we found a corruption rate of 40\% achieves the best VQA performance 
among $\{10\%, 20\%, ..., 80\%\}$, indicating an appropriate corruption rate is essential to control the task difficulty. However, the convergence rate of MLM is slow, making larger training steps necessary to achieve decent performance. Our method FLM surpasses MLM on NLVR$^2$ and image captioning, also showing an impressive 2.5\x speed-up of training time.

In Table~\ref{tab:datascale}, we show that the superiority of the proposed FLM consistently holds with a larger data scale or with more powerful visual features. 
The reason may be that FLM learns bidirectional context patterns by encoding the text sequence once and densely predicting the 100\% input in parallel, while MLM usually needs more pretraining steps to see such diverse patterns. 
% We conjecture that enlarging the pretraining data may amplify such damage of inefficient data utilization. 
Therefore, FLM is a friendly pretext task for accelerating training under low-resource scenarios, which to some extent, enjoys the high efficiency of AR/PrefixLM and the high performance of MLM.

As for underperformed retrieval performance compared with MLM, we conjecture that FLM with span corruptions has fewer corruption variants than MLM with random corruptions but focuses more on local semantics, which favors fine-grained tasks like VQA/captioning more than retrieval.

\subsection{Ablation Studies}
\label{subsec:label}
\paragraph{FLM Loss.}
The ablation study for the loss terms in FLM is shown in Table~\ref{tab:auxLoss}. With merely reconstruction loss $\mathcal{L}_R$, our model achieves better performance~(73.04 on VQA) compared with AR~(72.85) or PrefixLM~(72.64). When further introducing left-to-right or right-to-left intermediate caption loss, the model gains consistent improvements over two downstream tasks. Note that left-to-right loss shows non-trivial superiority to the right-to-left one, verifying the effectiveness of the causal relationships between words. By combing bidirectional caption loss, the model achieves 0.81/1.45 absolute gains over the model with merely reconstruction loss. 

\vspace{-0.5em}
\paragraph{Parameter Sharing.} During decomposed bidirectional encoding, 
parameter sharing is used in the text transformer for two unimodal encodings with different attention masks. Table~\ref{tab:sharing} shows that the shared text transformer clearly surpasses the unshared one, indicating that the two unidirectional representations could implicitly benefit each other by sharing the same feature space. 

\vspace{-0.5em}
\paragraph{Number of Reconstruction Layers.}
The reconstructor aims to construct several corrupted sequences upon the high-level representations and reconstruct the corrupted information in parallel. Table~\ref{tab:layerNum} shows that a deep structure of the reconstructor helps the downstream tasks. The multi-layer reconstructor gathers text and multimodal features from low to high levels, promising to enhance the representation ability.

\begin{table*}[!t]
\footnotesize
\centering
% \tablestyle{8pt}{1.0}
% \def\w{10pt} 
\scalebox{1.0}{
\setlength{\tabcolsep}{1.2 mm}{
  \begin{tabular}{llccccccccc}
    \toprule
    \multirow{2}{*}{\bf Model} & \multirow{2}{*}{\bf Pretrain. Task} & {\bf Pretrain. Time} &  \multicolumn{2}{c}{\bf VQAv2} & \multicolumn{2}{c}{\bf NLVR$^2$} &  \multicolumn{4}{c}{\bf COCO Captioning} \\
     & & (GPU Days) & \bf test-dev & \bf test-std & \bf dev & \bf test & \bf BLEU\@4 & \bf METEOR & \bf CIDEr & \bf SPICE \\
    % \shline

\midrule
% ####################### <10M ##############
\multicolumn{3}{l}{ { \it{\textbf{Pre-trained with $<$10M images}} } }\\
    % \hline
  UNITER$_{\text{LARGE}}$  ~\cite{chen2020uniter}  & MLM, ITM, MVM, WRA & 152 (V100) & 73.82 & 74.02  & 79.12 & 79.98 & - & -  \\
  UNIMO$_{\text{LARGE}}$~\cite{li2020unimo} & MLM, MVM, ITC & 640 (V100) & 75.06 & 75.27  & - & - &  - & {-}   \\
  OSCAR & MLM, ITM & 220 (V100) & 73.61 & 73.82 & 79.12 & 80.37 & 37.4 & \textbf{30.7} & 127.8 & 23.5 \\
  VinVL$_{\text{BASE}}$~\cite{zhang2021vinvl} & MLM, ITM & 320 (V100) & { 75.95}& 76.12 &  82.05 &  {83.08} & 38.2 & 30.3& 129.3 &23.6 \\
VinVL$_{\text{LARGE}}$~\cite{zhang2021vinvl} & MLM, ITM & 320 (V100) & { 76.52 }& 76.60 &  \underline{82.67} &  \textbf{83.98} & 38.5 & \underline{30.4} & {130.8} & 23.4 \\
  % PixelBERT~\cite{huang2020pixel} & MLM, ITM & - & 74.45 & 74.55 & 76.5 & 77.2 & - & - & -   \\
  CLIP-ViL~\cite{shen2021much} & MLM, ITM, VQA & 40 (A100) & 76.48 & { 76.70} & - & - & \demph{40.2$^*$} & \demph{29.7$^*$} & \demph{134.2$^*$} & \demph{23.8$^*$} \\
   
  ViLT~\cite{zhang2021vinvl} & MLM, ITM, WRA& 192 (V100)  &71.26 & - & 75.70 & 76.13 & - & - & \\
  ALBEF (4M)~\cite{li2021align} & MLM, ITM & 28 (A100) & 71.40 & - & - & 77.51 & - & - \\
  ALBEF (4M)~\cite{li2021align} & MLM, ITM, ITC & 28 (A100) & 74.54 & 74.70 & 80.24 & 80.50 & - & - \\
  
  METER$_{\text{BASE}}$~\cite{dou2022empirical}  & MLM, ITM & 64 (A100) & 77.68 &  77.64 & {82.33} & {83.05} & \underline{38.8} &30.0 &128.2 &23.0 \\

\rowcolor{Gray}
{Ours$_{\text{LARGE}}$ (4M)} & FLM &  18 (V100) & \underline{77.80} & \underline{77.84} & 81.77 & 81.83 & 38.3 & 30.2 & \underline{130.9} & -\\
\midrule
\multicolumn{3}{l}{ { \it{\textbf{Pre-trained with 10M$\sim$100M images}} } }\\
ALBEF (14M)~\cite{li2021align} & MLM, ITM, ITC &  140 (A100) & 75.84 & 76.04 & 82.55 & {83.14} & - & - \\
BLIP (14M)~\cite{li2022blip} & AR, ITM, ITC & 112 (A100)& {77.54} & {77.62} & {82.67} & 82.30 & {38.6} & - & {129.7} & - \\
\rowcolor{Gray}
{Ours$_{\text{LARGE}}$ (13M)} & FLM & {\color{black} 75 (V100)} & \textbf{78.18} & \textbf{78.24} & \textbf{82.90} & \underline{83.86}  & \textbf{39.1} & 30.3 & \textbf{132.7} & -\\
    \midrule
  \multicolumn{3}{l}{ \demph{ \it{\textbf{Pre-trained with $>$100M images}}}}\\
    % \hline    
    \demph{SimVLM$_{\text{BASE}}$ (1.8B)~\cite{wang2021simvlm}} & \demph{PrefixLM} & \demph{-} & \demph{77.87} & \demph{78.14} & \demph{81.72} & \demph{81.77} & \demph{39.0} & \demph{32.9} & \demph{134.8} & \demph{24.0} \\
    \demph{SimVLM$_{\text{HUGE}}$ (1.8B)~\cite{wang2021simvlm}} & \demph{PrefixLM} & \demph{-} & \demph{80.03} & \demph{80.34} & \demph{84.53} & \demph{85.15} & \demph{40.6} & \demph{33.7} & \demph{143.3} & \demph{25.4} \\
    \demph{LEMON (400M)}& \demph{MLM} & \demph{-}& \demph{-} & \demph{-}  & \demph{-}  & \demph{-} & \demph{40.3} & \demph{30.2} & \demph{133.3} & \demph{23.3} \\
\bottomrule
  \end{tabular}
}
}
\vspace{-.5em}
\caption{Comparisons with models on visual question answering, visual reasoning, and image captioning tasks. The best scores are in \textbf{bold}, and the second best scores are in \underline{underlined}.  MVM, ITC, ITM, and WRA represent masked vision modeling, image-text contrast, image-text matching, and word-region alignment, respectively. Ours$_{\rm LARGE}$ is trained with 30k/100k steps on 4M/13M data, respectively. 
The pretraining time of compared methods is provided by the original paper or estimated from open-sourcing code.}
\label{tab:sota}
\vspace{-2mm}
\end{table*}
\begin{table*}[!t]
\tablestyle{1pt}{1.0}
\def\w{20pt} 
\scalebox{1.0}{
  \begin{tabular}{llc|cccccc|cccccc}
    \toprule
    %\toprule
    \multirow{2}{*}{\bf Model} & \multirow{2}{*}{\bf Pretrain. Task} & {\bf Pretrain. Time} & \multicolumn{6}{c|}{\bf Flickr30k} & \multicolumn{6}{c}{\bf COCO}  \\
     & & (GPU Days) & \bf IR@1  & \bf IR@5  & \bf IR@10 & \bf TR@1 & \bf TR@5 & \bf TR@10 &  \bf IR@1  & \bf IR@5  & \bf IR@10 & \bf TR@1 & \bf TR@5 & \bf TR@10 \\
    \midrule
    \multicolumn{10}{l}{ { \it{\textbf{Pre-trained with $<$10M images}} } }\\
    UNITER$_{\text{LARGE}}$~\cite{chen2020uniter} & MLM, ITM, MVM, WRA & 152 (V100) & 75.56 & 94.08 & 96.76 & 87.30 & 98.00 & 99.20 & 52.93 & 79.93 & 87.95 & 65.68 & 88.56 & 93.76 \\
    UNIMO$_{\text{LARGE}}$~\cite{li2020unimo} &MLM, MVM, ITC&640 (V100)& 78.04 & 94.24 & 97.12 & 89.40 & 98.90 & \underline{ 99.80} & - & - & - & - & - & - \\
  VinVL$_{\text{LARGE}}$~\cite{zhang2021vinvl} &MLM, ITM& 320 (V100) & - & - & - & - & - & -  & \bf 58.8 & \bf 83.5 & \bf 90.3 & \underline{ 75.4} & \underline{ 92.9} & \underline{ 96.2} \\
  PixelBERT~\cite{huang2020pixel} &MLM, ITM & - & 71.5 & 92.1 & 95.8 & 87.0 & 98.9 & 99.5 & 50.1 & 77.6 & 86.2 & 63.6 & 87.5 & 93.6 \\
  ViLT~\cite{zhang2021vinvl} & MLM, ITM, WRA & 192 (V100)  & 64.4 & 88.7 & 93.8 & 83.5 & 96.7 & 98.6 & 42.7 & 72.9 & 83.1 & 61.5 & 86.3 & 92.7 \\
  ALBEF (4M)~\cite{li2021align} &MLM, ITM, ITC &28 (A100)& \underline{82.8} & \underline{96.7} & \underline{98.4} &  \underline{94.3} & { 99.4} &{ 99.8}  & 56.8 & 81.5 & 89.2 & 73.1 & 91.4 & 96.0\\
  METER$_{\text{BASE}}$~\cite{dou2022empirical} & MLM, ITM &64 (A100)& {82.22} &  {96.34} &  98.36 & \underline{94.30} & \bf 99.60 & \bf 99.90 & \underline{ 57.08} & \underline{ 82.66} & \underline{90.07} & \bf 76.16 & \bf 93.16 & \bf 96.82 \\
  \rowcolor{Gray}
  Ours$_{\text{LARGE}}$ (4M) &FLM& 18 (V100)&  74.53 & 93.96 & 97.26 & 88.10 & 98.30 & 99.60 & 46.46 & 75.43 & 85.09 & 62.84 & 86.64 & 93.00 \\
  \rowcolor{Gray}
  Ours$_{\text{LARGE}}$ (4M) &FLM, ITM& 57 (V100)& \textbf{83.40} & \textbf{97.04} & \textbf{98.72} & \bf 95.00 & \underline{99.50} & \bf 99.90 & 56.55 & 82.02 & 89.63 & 73.52 & 91.95 & 95.97 \\

  %########################## >10M images ##################
    \midrule
    \multicolumn{10}{l}{ \demph{ \it{\textbf{Pre-trained with $>$10M images}} } }\\
    \demph{ALBEF (14M)~\cite{li2021align}} & \demph{MLM, ITM, ITC} & \demph{60 (A100)} & \demph{85.6} & \demph{97.5} & \demph{98.9} & \demph{95.9} & \demph{99.8} & \demph{100.0} & \demph{60.7} & \demph{84.3} & \demph{90.5} & \demph{77.6} & \demph{94.3} & \demph{97.2}\\
    \demph{BLIP (14M)} & \demph{AR, ITM, ITC} & \demph{112 (A100)}  & \demph{87.2} & \demph{97.5} & \demph{98.8} & \demph{96.6} & \demph{99.8} & \demph{100.0} & \demph{63.1} & \demph{85.3} & \demph{91.1}& \demph{80.6} & \demph{95.2} & \demph{97.6}  \\
  \bottomrule
  \end{tabular}
  }
  \vspace{-.5em}
  \caption{Performance comparisons with models pre-trained on Flickr30k and COCO image retrieval (IR) and text retrieval (TR) tasks in the finetuning setting. The best scores are in \textbf{bold}, and the second best scores are in \underline{underlined}. }
  \label{tab:results2}
   \vspace{-4mm}
\end{table*}

\vspace{-0.5em}
\paragraph{Prediction Rate.}
We test the converged performance of the pretrained model with different prediction rates. To this end, we randomly mask a subset of output tokens from loss calculation. 
As shown in Table~\ref{tab:ablpr},  a lower prediction rate tends to achieve poor performance both on VQA and NLVR$^2$, probably suggesting that the prediction loss containing a larger number of tokens helps the optimization.

\vspace{-0.5em}
\paragraph{Corruption Rate.}
The corruption rate determines how much context should be used for predicting the corrupted tokens. It controls the difficulty of the reconstruction problem and closely affects model performance. We study the influence of corruption strategies in FLM. 
As shown in Fig.~\ref{tab:ablcr}, 
{First, we test the length of span corruption.} With the growth of span length, the VQA and NLVR$^2$ performance steadily reach their maximum values at the 30\%$\sim$40\% corruption rate. Our method keeps a 100\% prediction rate while allowing a customized corruption rate, which is hopeful to serve as a replacement for the widely-used MLM to improve convergence speed. 

Besides the span corruption which occurs after feature encoding, we also test the influence of pre-encoding corruption. We assign random corruptions to each token of the input sequence and then perform FLM to reconstruct all input tokens. With a 15\% corruption rate, random corruption could slightly increase the VQA score. But unfortunately, the NLVR$^2$ hurts with a larger corruption rate. We found that the optimal corruption rate may differ for different corruption methods. How to effectively fuse different types of corruption may be a promising direction to increase the diversity of contexts further.

\subsection{Comparsion with State-of-the-Arts}
The comparisons on VQA, NLVR$^2$ and image captioning are shown in Table~\ref{tab:sota}, without using complicated pretraining tasks like ITM and ITC, our method achieves competitive performance by merely using FLM as the pretraining task. Compared with prior arts, our method has appealing advantages regarding pretraining time: First, the proposed FLM helps the convergence speed by enabling 100\% token prediction. Second, we leverage FLM as the single pretraining objective, without relying on additional time-consuming pretraining objectives like ITM. 
Third, we use the patch-level image features instead of a heavy object detection used in~\cite{li2020oscar, zhang2021vinvl}.

The performance on cross-modal retrieval is shown in Table~\ref{tab:results2}. Our FLM-trained model performs poorly if directly fine-tuned on target downstream datasets.  Note that retrieval is heavily required for cross-modal alignment learning (\eg, ITM or ITC) on large-scale datasets since negative samples are essential to learning discriminative features. Therefore, we jointly use ITM and FLM to conduct pretraining to facilitate cross-modal alignments. By doing so, we obtain considerable performance gain and reach superior performance on Flickr30K and competitive performance on COCO over prior arts, suggesting the complementarity of FLM and other alignment-oriented objectives.

\section{Conclusion}
\vspace{-0.5em}
In this paper, we propose free language modeling (FLM), a new pretraining objective for accelerating vision-language pretraining. 
Different from previous language modeling methods, such as MLM and AR, FLM seamlessly disentangles the prediction rate from the tie-up with the corruption rate, meanwhile allowing a flexible corruption pattern for each prediction target. 
Experiments verify the effectiveness of the proposed FLM both in accuracy and efficiency. Our model could converge faster with a decent reduction of training time compared to MLM, while achieving comparable performance on multiple multimodal downstream tasks.

\vspace{-.3em}
\paragraph{Acknowledgement.}
This paper is partially supported by the National Key R\&D Program of China No. 2022ZD0161000, the General Research Fund of HK No.17200622, the National Natural Science Foundation of China under Grant No. 62122035 and 61972188. We thank Chengyue Wu for his technical assistance, and Jianhui Xu, Zhengxiao Du, and Zhichao Lu for their helpful comments.

{\small
\bibliographystyle{ieee_fullname}
\bibliography{egbib}

\begin{thebibliography}{10}\itemsep=-1pt

\bibitem{antol2015vqa}
Stanislaw Antol, Aishwarya Agrawal, Jiasen Lu, Margaret Mitchell, Dhruv Batra,
  C Lawrence~Zitnick, and Devi Parikh.
\newblock {VQA}: Visual question answering.
\newblock In {\em International Conference on Computer Vision (ICCV)}, 2015.

\bibitem{bao2022vl}
Hangbo Bao, Wenhui Wang, Li Dong, and Furu Wei.
\newblock Vl-beit: Generative vision-language pretraining.
\newblock {\em arXiv preprint arXiv:2206.01127}, 2022.

\bibitem{bitton2021data}
Yonatan Bitton, Gabriel Stanovsky, Michael Elhadad, and Roy Schwartz.
\newblock Data efficient masked language modeling for vision and language.
\newblock {\em arXiv preprint arXiv:2109.02040}, 2021.

\bibitem{brown2020language}
Tom Brown, Benjamin Mann, Nick Ryder, Melanie Subbiah, Jared~D Kaplan, Prafulla
  Dhariwal, Arvind Neelakantan, Pranav Shyam, Girish Sastry, Amanda Askell,
  et~al.
\newblock Language models are few-shot learners.
\newblock {\em Advances in neural information processing systems},
  33:1877--1901, 2020.

\bibitem{byun2022grit}
Jaeseok Byun, Taebaek Hwang, Jianlong Fu, and Taesup Moon.
\newblock Grit-vlp: Grouped mini-batch sampling for efficient vision and
  language pre-training.
\newblock {\em arXiv preprint arXiv:2208.04060}, 2022.

\bibitem{changpinyo2021conceptual}
Soravit Changpinyo, Piyush Sharma, Nan Ding, and Radu Soricut.
\newblock Conceptual 12m: Pushing web-scale image-text pre-training to
  recognize long-tail visual concepts.
\newblock In {\em Conference on Computer Vision and Pattern Recognition
  (CVPR)}, 2021.

\bibitem{chen2020uniter}
Yen-Chun Chen, Linjie Li, Licheng Yu, Ahmed El~Kholy, Faisal Ahmed, Zhe Gan, Yu
  Cheng, and Jingjing Liu.
\newblock {UNITER}: Universal image-text representation learning.
\newblock In {\em European Conference on Computer Vision (ECCV)}, 2020.

\bibitem{chowdhery2022palm}
Aakanksha Chowdhery, Sharan Narang, Jacob Devlin, Maarten Bosma, Gaurav Mishra,
  Adam Roberts, Paul Barham, Hyung~Won Chung, Charles Sutton, Sebastian
  Gehrmann, et~al.
\newblock Palm: Scaling language modeling with pathways.
\newblock {\em arXiv preprint arXiv:2204.02311}, 2022.

\bibitem{devlin2018bert}
Jacob Devlin, Ming-Wei Chang, Kenton Lee, and Kristina Toutanova.
\newblock {BERT}: Pre-training of deep bidirectional transformers for language
  understanding.
\newblock In {\em Conference of the North {A}merican Chapter of the Association
  for Computational Linguistics (NAACL)}, 2019.

\bibitem{dou2022empirical}
Zi-Yi Dou, Yichong Xu, Zhe Gan, Jianfeng Wang, Shuohang Wang, Lijuan Wang,
  Chenguang Zhu, Pengchuan Zhang, Lu Yuan, Nanyun Peng, et~al.
\newblock An empirical study of training end-to-end vision-and-language
  transformers.
\newblock In {\em Proceedings of the IEEE/CVF Conference on Computer Vision and
  Pattern Recognition}, pages 18166--18176, 2022.

\bibitem{du2022glm}
Zhengxiao Du, Yujie Qian, Xiao Liu, Ming Ding, Jiezhong Qiu, Zhilin Yang, and
  Jie Tang.
\newblock Glm: General language model pretraining with autoregressive blank
  infilling.
\newblock In {\em Proceedings of the 60th Annual Meeting of the Association for
  Computational Linguistics (Volume 1: Long Papers)}, pages 320--335, 2022.

\bibitem{he2020deberta}
Pengcheng He, Xiaodong Liu, Jianfeng Gao, and Weizhu Chen.
\newblock De{BERT}a: Decoding-enhanced bert with disentangled attention.
\newblock {\em arXiv preprint}, 2020.

\bibitem{huang2020pixel}
Zhicheng Huang, Zhaoyang Zeng, Bei Liu, Dongmei Fu, and Jianlong Fu.
\newblock Pixel-{BERT}: Aligning image pixels with text by deep multi-modal
  transformers.
\newblock {\em arXiv preprint}, 2020.

\bibitem{jia2021scaling}
Chao Jia, Yinfei Yang, Ye Xia, Yi-Ting Chen, Zarana Parekh, Hieu Pham, Quoc~V
  Le, Yunhsuan Sung, Zhen Li, and Tom Duerig.
\newblock Scaling up visual and vision-language representation learning with
  noisy text supervision.
\newblock {\em arXiv preprint}, 2021.

\bibitem{kim2021vilt}
Wonjae Kim, Bokyung Son, and Ildoo Kim.
\newblock {ViLT}: Vision-and-language transformer without convolution or region
  supervision.
\newblock In {\em International Conference on Machine Learning (ICML)}, 2021.

\bibitem{krishna2017visual}
Ranjay Krishna, Yuke Zhu, Oliver Groth, Justin Johnson, Kenji Hata, Joshua
  Kravitz, Stephanie Chen, Yannis Kalantidis, Li-Jia Li, David~A Shamma, et~al.
\newblock Visual {G}enome: Connecting language and vision using crowdsourced
  dense image annotations.
\newblock {\em International Journal of Computer Vision (IJCV)}, 2017.

\bibitem{li2022blip}
Junnan Li, Dongxu Li, Caiming Xiong, and Steven Hoi.
\newblock Blip: Bootstrapping language-image pre-training for unified
  vision-language understanding and generation.
\newblock {\em arXiv preprint arXiv:2201.12086}, 2022.

\bibitem{li2021align}
Junnan Li, Ramprasaath~R Selvaraju, Akhilesh~Deepak Gotmare, Shafiq Joty,
  Caiming Xiong, and Steven Hoi.
\newblock Align before fuse: Vision and language representation learning with
  momentum distillation.
\newblock In {\em Conference on Neural Information Processing Systems
  (NeurIPS)}, 2021.

\bibitem{li2019visualbert}
Liunian~Harold Li, Mark Yatskar, Da Yin, Cho-Jui Hsieh, and Kai-Wei Chang.
\newblock Visual{BERT}: A simple and performant baseline for vision and
  language.
\newblock {\em arXiv preprint}, 2019.

\bibitem{li2022grounded}
Liunian~Harold Li, Pengchuan Zhang, Haotian Zhang, Jianwei Yang, Chunyuan Li,
  Yiwu Zhong, Lijuan Wang, Lu Yuan, Lei Zhang, Jenq-Neng Hwang, et~al.
\newblock Grounded language-image pre-training.
\newblock In {\em Proceedings of the IEEE/CVF Conference on Computer Vision and
  Pattern Recognition}, pages 10965--10975, 2022.

\bibitem{li2020unimo}
Wei Li, Can Gao, Guocheng Niu, Xinyan Xiao, Hao Liu, Jiachen Liu, Hua Wu, and
  Haifeng Wang.
\newblock Unimo: Towards unified-modal understanding and generation via
  cross-modal contrastive learning.
\newblock In {\em Annual Meeting of the Association for Computational
  Linguistics (ACL)}, 2021.

\bibitem{li2020oscar}
Xiujun Li, Xi Yin, Chunyuan Li, Pengchuan Zhang, Xiaowei Hu, Lei Zhang, Lijuan
  Wang, Houdong Hu, Li Dong, Furu Wei, et~al.
\newblock Oscar: Object-semantics aligned pre-training for vision-language
  tasks.
\newblock In {\em European Conference on Computer Vision (ECCV)}, 2020.

\bibitem{li2021supervision}
Yangguang Li, Feng Liang, Lichen Zhao, Yufeng Cui, Wanli Ouyang, Jing Shao,
  Fengwei Yu, and Junjie Yan.
\newblock Supervision exists everywhere: A data efficient contrastive
  language-image pre-training paradigm.
\newblock {\em arXiv preprint arXiv:2110.05208}, 2021.

\bibitem{lin2014microsoft}
Tsung-Yi Lin, Michael Maire, Serge Belongie, James Hays, Pietro Perona, Deva
  Ramanan, Piotr Doll{\'a}r, and C~Lawrence Zitnick.
\newblock Microsoft {COCO}: Common objects in context.
\newblock In {\em European Conference on Computer Vision (ECCV)}, 2014.

\bibitem{liu2019roberta}
Yinhan Liu, Myle Ott, Naman Goyal, Jingfei Du, Mandar Joshi, Danqi Chen, Omer
  Levy, Mike Lewis, Luke Zettlemoyer, and Veselin Stoyanov.
\newblock Ro{BERT}a: A robustly optimized bert pretraining approach.
\newblock {\em arXiv preprint}, 2019.

\bibitem{loshchilov2017decoupled}
Ilya Loshchilov and Frank Hutter.
\newblock Decoupled weight decay regularization.
\newblock {\em arXiv preprint arXiv:1711.05101}, 2017.

\bibitem{lu2019vilbert}
Jiasen Lu, Dhruv Batra, Devi Parikh, and Stefan Lee.
\newblock Vilbert: Pretraining task-agnostic visiolinguistic representations
  for vision-and-language tasks.
\newblock In {\em Conference on Neural Information Processing Systems
  (NeurIPS)}, 2019.

\bibitem{ordonez2011im2text}
Vicente Ordonez, Girish Kulkarni, and Tamara Berg.
\newblock Im2text: Describing images using 1 million captioned photographs.
\newblock In {\em Conference on Neural Information Processing Systems
  (NeurIPS)}, 2011.

\bibitem{radford2021learning}
Alec Radford, Jong~Wook Kim, Chris Hallacy, Aditya Ramesh, Gabriel Goh,
  Sandhini Agarwal, Girish Sastry, Amanda Askell, Pamela Mishkin, Jack Clark,
  et~al.
\newblock Learning transferable visual models from natural language
  supervision.
\newblock In {\em International Conference on Machine Learning (ICML)}, 2021.

\bibitem{sharma2018conceptual}
Piyush Sharma, Nan Ding, Sebastian Goodman, and Radu Soricut.
\newblock Conceptual captions: A cleaned, hypernymed, image alt-text dataset
  for automatic image captioning.
\newblock In {\em Annual Meeting of the Association for Computational
  Linguistics (ACL)}, 2018.

\bibitem{shen2021much}
Sheng Shen, Liunian~Harold Li, Hao Tan, Mohit Bansal, Anna Rohrbach, Kai-Wei
  Chang, Zhewei Yao, and Kurt Keutzer.
\newblock How much can clip benefit vision-and-language tasks?
\newblock {\em arXiv preprint}, 2021.

\bibitem{suhr2018corpus}
Alane Suhr, Stephanie Zhou, Ally Zhang, Iris Zhang, Huajun Bai, and Yoav Artzi.
\newblock A corpus for reasoning about natural language grounded in
  photographs.
\newblock In {\em Annual Meeting of the Association for Computational
  Linguistics (ACL)}, 2019.

\bibitem{tan-bansal-2019-lxmert}
Hao Tan and Mohit Bansal.
\newblock {LXMERT}: Learning cross-modality encoder representations from
  transformers.
\newblock In {\em Conference on Empirical Methods in Natural Language
  Processing (EMNLP)}, 2019.

\bibitem{vaswani2017attention}
Ashish Vaswani, Noam Shazeer, Niki Parmar, Jakob Uszkoreit, Llion Jones,
  Aidan~N Gomez, {\L}ukasz Kaiser, and Illia Polosukhin.
\newblock Attention is all you need.
\newblock In {\em Conference on Neural Information Processing Systems
  (NeurIPS)}, 2017.

\bibitem{wang2022vlmixer}
Teng Wang, Wenhao Jiang, Zhichao Lu, Feng Zheng, Ran Cheng, Chengguo Yin, and
  Ping Luo.
\newblock Vlmixer: Unpaired vision-language pre-training via cross-modal
  cutmix.
\newblock In {\em International Conference on Machine Learning}, pages
  22680--22690. PMLR, 2022.

\bibitem{wang2022image}
Wenhui Wang, Hangbo Bao, Li Dong, Johan Bjorck, Zhiliang Peng, Qiang Liu, Kriti
  Aggarwal, Owais~Khan Mohammed, Saksham Singhal, Subhojit Som, et~al.
\newblock Image as a foreign language: Beit pretraining for all vision and
  vision-language tasks.
\newblock {\em arXiv preprint arXiv:2208.10442}, 2022.

\bibitem{wang2021simvlm}
Zirui Wang, Jiahui Yu, Adams~Wei Yu, Zihang Dai, Yulia Tsvetkov, and Yuan Cao.
\newblock Simvlm: Simple visual language model pretraining with weak
  supervision.
\newblock {\em arXiv preprint}, 2021.

\bibitem{wettig2022should}
Alexander Wettig, Tianyu Gao, Zexuan Zhong, and Danqi Chen.
\newblock Should you mask 15\% in masked language modeling?
\newblock {\em arXiv preprint arXiv:2202.08005}, 2022.

\bibitem{yang2019xlnet}
Zhilin Yang, Zihang Dai, Yiming Yang, Jaime Carbonell, Russ~R Salakhutdinov,
  and Quoc~V Le.
\newblock Xlnet: Generalized autoregressive pretraining for language
  understanding.
\newblock {\em Advances in neural information processing systems}, 32, 2019.

\bibitem{yao2021filip}
Lewei Yao, Runhui Huang, Lu Hou, Guansong Lu, Minzhe Niu, Hang Xu, Xiaodan
  Liang, Zhenguo Li, Xin Jiang, and Chunjing Xu.
\newblock Filip: Fine-grained interactive language-image pre-training.
\newblock {\em arXiv preprint arXiv:2111.07783}, 2021.

\bibitem{yu2022coca}
Jiahui Yu, Zirui Wang, Vijay Vasudevan, Legg Yeung, Mojtaba Seyedhosseini, and
  Yonghui Wu.
\newblock Coca: Contrastive captioners are image-text foundation models.
\newblock {\em arXiv preprint arXiv:2205.01917}, 2022.

\bibitem{zeng2021multi}
Yan Zeng, Xinsong Zhang, and Hang Li.
\newblock Multi-grained vision language pre-training: Aligning texts with
  visual concepts.
\newblock {\em arXiv preprint arXiv:2111.08276}, 2021.

\bibitem{zhang2021vinvl}
Pengchuan Zhang, Xiujun Li, Xiaowei Hu, Jianwei Yang, Lei Zhang, Lijuan Wang,
  Yejin Choi, and Jianfeng Gao.
\newblock {VinVL}: Revisiting visual representations in vision-language models.
\newblock In {\em Conference on Computer Vision and Pattern Recognition
  (CVPR)}, 2021.

\end{thebibliography}
}

\newpage
\section{Supplementary Materials}

% \subsection{Pretraining Hyper-parameters }

% overall table of all ablations
\begin{table*}[t]
% \vspace{-.2em}
\centering
\small
%#################################################
% Loss term
%#################################################
\subfloat[
\textbf{4M data}
    \label{tab:hparam4m}
]{
% \centering
\begin{minipage}{0.4\linewidth}{\begin{center}
\renewcommand\arraystretch{1.0}
\setlength{\tabcolsep}{1.0 mm}{
    \begin{tabular}{lcc}
        & Ours$_{\rm BASE}$ & Ours$_{\rm LARGE}$ \\
        \midrule
        patch size & 32/16 & 14 \\
        image size & 288\x288 & 336\x336 \\
        learning rate & 4e-4 & 4e-4 \\
        learning rate (pretrained layers) & 8e-5 & 8e-5 \\
        warmup rate & 0.05 & 0.05 \\
        training steps & 30k & 30k
    \end{tabular}}
\end{center}}\end{minipage}
}
\hspace{3em}
%#################################################
% Parameter sharing
%#################################################
\subfloat[
\textbf{13M data}
        \label{tab:hparam13m}
]{
\begin{minipage}{0.4\linewidth}{\begin{center}
\renewcommand\arraystretch{1.0}
\setlength{\tabcolsep}{1.0 mm}{
    \begin{tabular}{lcc}
        & Ours$_{\rm BASE}$ & Ours$_{\rm LARGE}$ \\
        \midrule
        patch size & 32/16 & 14 \\
        image size & 288\x288 & 224\x224 \\
        learning rate & 4e-4 & 4e-4 \\
        learning rate (pretrained layers) & 8e-5 & 8e-5 \\
        warmup rate & 0.05 & 0.05 \\
        training steps & 30k & 100k \\
    \end{tabular}}
\end{center}}\end{minipage}
}
\hspace{2em}
\caption{Hyper-parameters for pretraining.}
\label{tab:hparam}
\end{table*}

\subsection{Implementation Details}

We list the hyperparameters for pretraining in Table~\ref{tab:hparam}. The implementation details for downstream tasks are described as follows.

% \vspace{-.3em}
\paragraph{Visual Question Answering (VQA).} We follow~\cite{chen2020uniter} to consider VQA as a classification problem on 3129 most frequent answers. We input a single $\rm [CLS]$ token upon the reconstructor and regard its output representation as the multimodal features, followed by an MLP classifier to obtain the final classification probability. 
Following~\cite{dou2022empirical}, during fine-tuning, the learning rates of the image encoder and bottom layers of the text transformer are 5e-6, and those for top layers of the text transformer and the reconstructor are 2.5e-5. 

% \vspace{-.3em}
\paragraph{Natural Language for Visual Reasoning for Real (NLVR$^2$).} The task aims to distinguish whether the natural language description is true given a pair of images. We follow~\cite{chen2020uniter} to consider the input triplet (a sentence and two images) as two image-text pairs. For each image-text pair, we obtain the $\rm [CLS]$ embedding from the reconstructor as the multimodal embeddings. The two embeddings are concatenated and input into an MLP for binary classification. Following~\cite{dou2022empirical}, during fine-tuning, the learning rates of the image encoder and bottom layers of the text transformer and the reconstructor are set to 1e-5, and those for top layers of the text transformer are 5e-5. 

% \vspace{-.3em}
\paragraph{Image Retrieval (IR) and Text Retrieval (TR).}  We use the image-text matching loss to finetune the pretrained model on the downstream retrieval datasets, \ie, COCO and Flickr30K. During training, we construct random negative pairs by replacing the paired images with random images sampled from the dataset. An MLP is applied on the $\rm [CLS]$ embedding of the reconstructor for binary classification. The learning rates of the image encoder and bottom layers of the text transformer are 5e-6, and those for top layers and the reconstructor of the text transformer are 2.5e-5. 

% \vspace{-.3em}
\paragraph{Image Captioning.} Since the intermediate loss of our model considers an autoregressive generation process, the finetuning performance or zero-shot performance (shown in Sec.~\ref{sec:zscaption}) on captioning datasets could be evaluated. For finetuning performance, we remove the reconstructor and finetune the model with unidirectional captioning loss. Note that we do not use beam search for simplicity. The learning rates of the image encoder and bottom layers of the text transformer are set to 3e-6, and those for top layers of the text transformer and the reconstructor are 1.5e-5.

\subsection{Additional Analysis}
\label{sec:zscaption}

\paragraph{More Evidence of the Motivation.}
\begin{table}[]
\small
\center
    \renewcommand\arraystretch{1.0}
    \setlength{\tabcolsep}{2.5 mm}{
    \begin{tabular}{lcccc}
    \toprule
    \multirow{2}{*}{Model} & \multirow{2}{*}{$r_{\rm corr}$} & \multirow{2}{*}{$r_{\rm pred}$} & \multicolumn{2}{c}{MLM loss} \\
    &&&50\% steps & 100\% steps \\
    \midrule
       ViLT & 40\% & 20\% & 2.161 & 2.044 \\
       ViLT & 40\% & 40\% & 1.872 & 1.769\\
       ViLT & 15\% & 7.5\% &  1.808 & 1.655 \\
       ViLT & 15\% & 15\% &  1.699 &  1.574 \\
       \midrule
       RoBERTa & 40\% & 20\% & 3.857 & 3.501 \\
       RoBERTa & 40\% & 40\% & 3.641 & 3.371 \\
    \bottomrule
    \end{tabular}
    }
    % \captionsetup{font={scriptsize,stretch=0.85}}
    \caption{Varying prediction and corruption rates. 
    For ViLT~\cite{kim2021vilt}, we follow the official recipe with 25k steps. For RoBERTa~\cite{liu2019roberta}, we use an efficient recipe with 23k steps from~\cite{wettig2022should}.
    }
    \label{tab:vilt}
\end{table}
In Table~\ref{tab:vilt}, 
our key motivation -- limited prediction rate impedes convergence speed -- is justified in wider environments with different VLP structures and pretraining data, \ie, ViLT (single-encoder structure) and RoBERTa (pretrained on text-only datasets). A consistent trend is found that lower prediction rates gain higher MLM losses, verifying that such motivation is reasonable among different structures and datasets, even in text-only pretraining.

\paragraph{MLM with Varied Masking Ratios.}
We explore how much acceleration the MLM-based methods could achieve with a larger mask ratio. As shown in Table~\ref{tab:mlmmaskrate}, when increasing the mask ratio from 0.2 to 0.8, a mask ratio of 0.6 achieves the best performance within the 30k steps. However, when the training steps grow after 50k steps, a mask rate of 0.4 achieves the best. Compared with 0.6, MLM with a 0.8 mask ratio shows slower convergence, probably caused by that larger corruption rate increasing the learning difficulty. We conclude that for MLM, the corruption rate and prediction rate are tied-up by the mask ratio, and a proper corruption rate is
achieved at the cost of a large portion of output tokens being excluded from prediction loss.

\paragraph{PrefixLM with Varied $r_{\rm pred}$ and $r_{\rm corr}$.} Similarly to MLM, the
corruption rate and prediction rate in PrefixLM are tied-up in nature. We found all experiments have a similar converge rate but much different converged performance. Table~\ref{tab:prefixLM} shows the best result is achieved with  $r_{\rm pred}$ = 2$ \cdot r_{\rm corr}\!$ = 75\%. 
Lower $r_{\rm corr}$ results in easier tasks and lower representability, while a much larger one may cause learning collapse.

\begin{table}[]
    \centering
    \small
    \setlength{\tabcolsep}{1.0 mm}{
    \begin{tabular}{lc|cccccc}
    \toprule
    \multirow{2}{*}{Method} & \multirow{2}{*}{Mask Ratio} & \multicolumn{6}{c}{Training Steps} \\
    & & 10k & 20k & 30k & 50k & 80k & 100k \\
    \midrule
        MLM & 0.2 & 51.07 & 74.38 & 76.83 & 77.65 & 78.20  &  78.21 \\
        MLM & 0.4 & 51.69 & 76.14 & 77.59 & \textbf{78.46} & \textbf{78.64} & \textbf{78.30} \\
        MLM & 0.6 & \textbf{62.62} & \textbf{76.69} & \textbf{78.01} & 77.97 & 78.33 & 77.91 \\
        MLM & 0.8 & 51.07 & 76.32 & 77.24 & 78.04 & 77.78 & 78.20 \\
\bottomrule
    \end{tabular}
    }
    \caption{NLVR$^2$ performance of different mask ratios in MLM. All models are trained with a maximum of 100k steps on 4M data.}
    \label{tab:mlmmaskrate}
\end{table}

\begin{table}[]
\small
\center
    \renewcommand\arraystretch{1.0}
    \setlength{\tabcolsep}{2.5 mm}{
    \begin{tabular}{cccc}
    \toprule
    Method & $r_{\rm corr}$ & $r_{\rm pred}$ & NLVR$^2$ \\
        \midrule
       PrefixLM & 12.5\% & 25\% &  75.00 \\
       PrefixLM & 25.0\% & 50\% &  76.17 \\
       PrefixLM & 37.5\% & 75\% &   76.78 \\
       PrefixLM & 50.0\% & 100\% &  76.29 \\
\bottomrule
    \end{tabular}
    }
    \caption{Varying $r_{\rm corr}$ and $r_{\rm pred}$ of PrefixLM. It is achieved by modifying the distribution of prefix length. All models are trained with a maximum of 30k steps on 4M data.
    }
    \label{tab:prefixLM}
\end{table}

\begin{table}[]
    \centering
    \small
    \setlength{\tabcolsep}{2.2 mm}{
    \begin{tabular}{l|cccccc}
    \toprule
    \multirow{2}{*}{Method} & \multicolumn{3}{c}{Zero-shot  Captioning} & \multicolumn{3}{c}{Finetuned Captioning} \\
    & B@4 & M & C & B@4 & M & C \\
    \midrule
        AR & \textbf{24.9} & \textbf{21.6} & \textbf{80.3} & 35.70 & 28.86 & 120.6  \\
        PrefixLM &  22.6 & 20.2 & 73.3 & 35.50 & 28.79 & 119.4 \\
        FLM & 20.7 & 19.6 & 70.3  & \textbf{36.68} & \textbf{29.17} & \textbf{123.0} \\
\bottomrule
    \end{tabular}
    }
    \caption{Image captioning performance of different pretraining objectives on COCO. B@4, M, C are short for BLEU@4, METEOR, CIDEr, respectively.}
    \label{tab:zscap}
\end{table}

\paragraph{Zero-shot Captioning Performance.} 
The proposed FLM objectives include two parts, a reconstruction loss for solving corruption-prediction tasks with bidirectional contexts, and an intermediate loss that supervises the model and focuses more on temporal relationships with unidirectional context. 
% The intermediate can  be considered as the captioning loss. 
After pretraining, we could directly test the zero-shot captioning performance without further finetuning on target datasets. The zero-shot performance of different pretraining objectives is shown in Table~\ref{tab:zscap}. While MLM-based methods can not be directly used in captioning tasks, FLM achieves reasonable zero-shot captioning performance. However, for finetuned captioning performance, FLM achieves better performance than AR/PrefixLM. We conjecture that the FLM objectives could capture more generalizable features than AR/PrefixLM for captioning after finetuning.

\end{document}